\documentclass{article}

\usepackage{microtype}
\usepackage{graphicx}
\usepackage{subfigure}
\usepackage{booktabs} 

\usepackage{hyperref}

\usepackage[accepted]{icml2021}

\icmltitlerunning{Short-term Hourly Streamflow Prediction with Graph Convolutional GRU Networks}

\begin{document}

\twocolumn[
\icmltitle{Short-term Hourly Streamflow Prediction with Graph Convolutional GRU Networks}




\begin{icmlauthorlist}
\icmlauthor{Muhammed Sit}{iihr}
\icmlauthor{Bekir Demiray}{iihr}
\icmlauthor{Ibrahim Demir}{iihr}

\end{icmlauthorlist}

\icmlaffiliation{iihr}{IIHR—Hydroscience \& Engineering, The University of Iowa, Iowa City, Iowa, USA}

\icmlcorrespondingauthor{Muhammed Sit}{muhammed-sit@uiowa.edu}

\icmlkeywords{Environment, Graph Neural Networks, Streamflow, Flood Precipitation}

\vskip 0.3in
]



\printAffiliationsAndNotice{}  

\begin{abstract}
The frequency and impact of floods are expected to increase due to climate change. It is crucial to predict streamflow, consequently flooding, in order to prepare and mitigate its consequences in terms of property damage and fatalities. This paper presents a Graph Convolutional GRUs based model to predict the next 36 hours of streamflow for a sensor location using the upstream river network. As shown in experiment results, the model presented in this study provides better performance than the persistence baseline and a Convolutional Bidirectional GRU network for the selected study area in short-term streamflow prediction.

\end{abstract}

\section{Introduction}
\label{intro}

The number and devastating impacts of natural disasters have grown significantly worldwide. The latest report released by Munich Re (2020) states that natural disasters such as hurricanes, floods, and other disasters caused more than \$210 billion estimated damage worldwide, while \$95 billion of the damage occurred in the US. According to the same report, summer floods in China were the costliest natural disaster worldwide in 2020, and the number of fatalities in floods was higher than other natural disasters \cite{munichre}. Recent studies show that the frequency and impact of flooding increases in certain regions as a result of climate change \cite{davenport2021contribution, ncei, tabari2020climate} due to an increase in sea level \cite{strauss2016unnatural} and frequency of extreme precipitation \cite{diffenbaugh2017quantifying}, or intensifying hurricane rainfall \cite{trenberth2018hurricane}. Thus, it is crucial to predict streamflow and, consequently, flooding to mitigate its devastating effects in terms of damage and fatalities. 

Many physical and data-driven methods have been proposed to achieve accurate streamflow predictions, and recent studies show that deep learning models often provide more accurate results than physical-based models \cite{gauch2019data, xiang2020distributed, xiang2021regional}. Recurrent neural networks based approaches are mainly used for the task as a result of the success of these architectures on time series problems \cite{sit2019decentralized}. \cite{hu2018deep} proposed a Long short-term memory (LSTM) \cite{hochreiter1997long} model that predicts the hourly streamflow from 1 to 6 hours lead time. \cite{kratzert2018rainfall} developed another LSTM model that predicts daily runoff for the first time. \cite{xiang2020distributed} developed a model that uses multiple GRUs  \cite{cho2014learning} and Time-distributed layers in order to predict up to 120 hours of streamflow. Many studies aim to predict streamflow based on various data such as evapotranspiration, current streamflow, or weather data. More detailed information about deep learning studies on streamflow prediction can be found in \cite{sit2020comprehensive}. 

Graph Neural Networks (GNNs) and variants including Graph Convolutional Networks (GCNs), Graph Attention Networks (GATs), or Graph Recurrent Networks (GRNs) have gained much attention as a result of their performance on many deep learning tasks comprising sequenced data points which can be expressed as a graph \cite{wu2020comprehensive, zhou2020graph}. For instance, \cite{seo2018structured} proposes a convolutional recurrent neural network architecture that captures spatial dependencies with CNNs and identifies dynamic patterns with GRUs in the structured data sequences by showing two use cases; predicting moving MNIST data and modeling natural language with the Penn Treebank dataset. \cite{bai2020adaptive} uses a graph convolutional recurrent network with two new modules that allow to learn node-specific patterns and discover spatial correlations from data separately for a traffic prediction task. Based on successful implementations of GNNs on structured sequences, in this paper, we present a model based on Graph Convolutional GRUs for streamflow prediction that we will refer to as StreamGConvGRU. Furthermore, we show our preliminary results for StreamGConvGRU using real-world data for a small network of streamflow sensors. To the best of our knowledge, this is the first work that presents at least some preliminary results for streamflow prediction using Graph Neural Networks. 

\section{Methodology}
GNNs expect the input data as a graph and successively outputs a graph. Since most rivers are connected to each other and form a network of rivers, locations along rivers, more specifically stream gauge locations along a river network, could be converted into a graph and fed to a GNN.

\begin{figure}[!htb]
\vskip 0.2in
\begin{center}
\centerline{\frame{\includegraphics[width=\columnwidth]{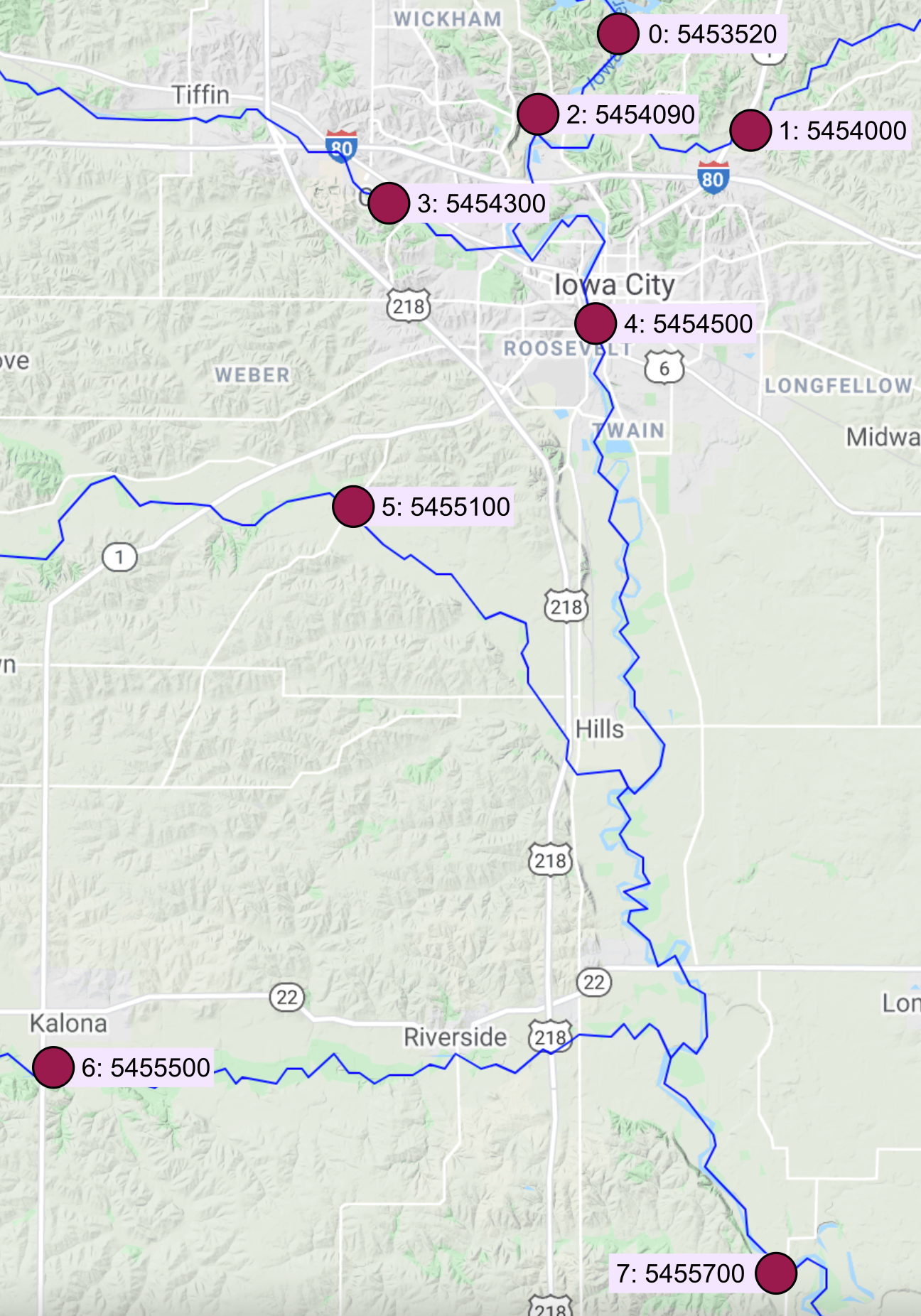}}}
\caption{Visualization of study area and USGS sensor locations on Google Maps with their \textit{id\_in\_graph: usgs\_id}}
\label{sensors}
\end{center}
\vskip -0.2in
\end{figure}

\subsection{Dataset}

The United States Geological Survey (USGS) maintains a network of stream gauges while actively deploying new ones all around the United States. The historical measurements for each of these sensors with a temporal resolution of 15 minutes are publicly available. Even though there are hundreds of USGS stream gauges in each watershed, in order to provide a proof of concept model setup, we chose to use eight sensors in Johnson County, Iowa, that form a small subnetwork. Seven of these stream gauges are within the watershed of the eighth one, and they feed the water to the eighth stream gauge located at the outlet of the watershed. In this study, we aim to predict the next 36 hours of measurements for the eighth stream gauge at the outlet by predominantly using 36 hours of historical streamflow data for all eight stream gauges. USGS sensor ids of these stream gauges are as follows, 05453520,  05454000, 05454090, 05454300, 05455100, 05454500, 05455500, 05455700, the last one being the sensor for that we aim to predict the future streamflow values (Figure \ref{sensors}).

Along with historic stream measurements, the StreamGConvGRU is also fed with precipitation measurements for the past 36 hours and the next 36 hours for all the stream gauge locations. The resource we employed for the rainfall data is the Stage IV hourly multi-sensor precipitation analysis. Since all the precipitation that falls into a river’s watershed affects the streamflow, the rainfall data within every sensor’s watershed were aggregated by summing up. Also, in order to match stream gauge temporal resolution with Stage IV, the USGS sensor measurements were aggregated to hourly by averaging them.

\begin{figure}[!htb]
\vskip 0.2in
\begin{center}
\centerline{\includegraphics[width=\columnwidth]{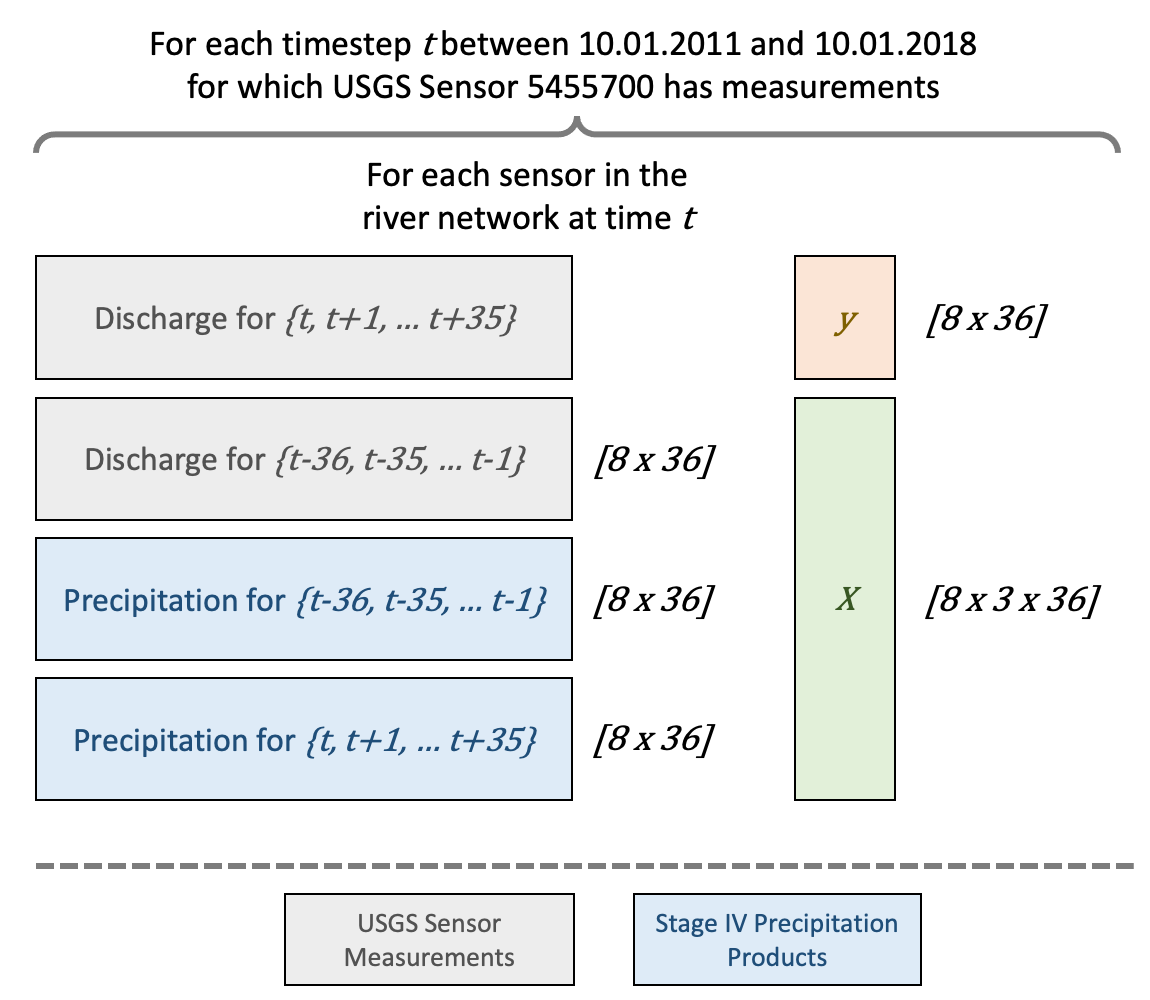}}
\caption{Summary of data sources and how each snapshot was formed.} 
\label{data}
\end{center}
\vskip -0.2in
\end{figure}

\begin{table}[!htb]
\caption{Dataset splits and, start and end dates for each split.}
\label{dataset-split}
\vskip 0.15in
\begin{center}
\begin{small}
\begin{sc}
\begin{tabular}{lcccr}
\toprule
Split & Start Date & End Date & \# of Snapshots \\
\midrule
Train    & 10.01.2011 & 10.01.2016 & 32789\\
Validation    & 10.01.2016 & 10.01.2017 & 6896\\
Test    & 10.01.2017 & 10.01.2018 & 8253 \\
\bottomrule
\end{tabular}
\end{sc}
\end{small}
\end{center}
\vskip -0.1in
\end{table}

\begin{figure*}[!ht]
\vskip 0.2in
\begin{center}
\centerline{\includegraphics[height=5cm,keepaspectratio]{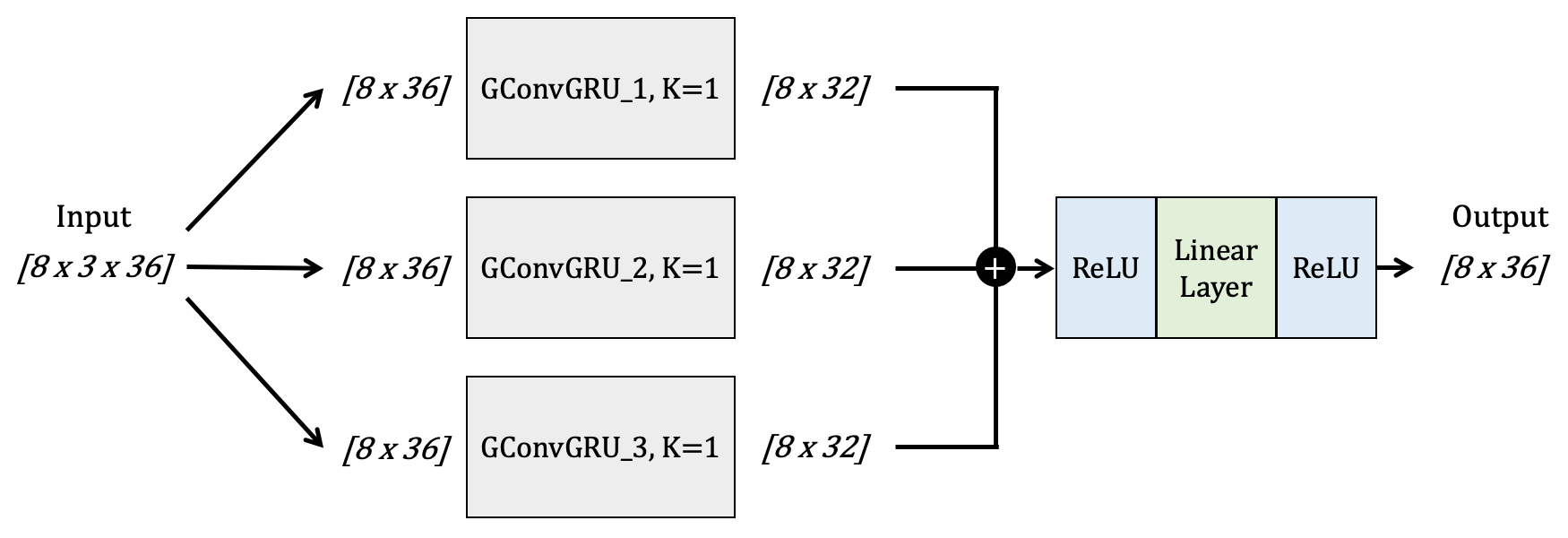}}
\caption{Architecture proposed for StreamGConvGRU.}
\label{arch}
\end{center}
\vskip -0.2in
\end{figure*}

Within the selected date range (10/1/2011 - 10/01/2018), a snapshot was created for each hour except for the times when the USGS sensor 05455700 is down. In order to ensure a continuous dataset and introduce noise to the training process, we simply used zeros for when any of the seven upstream sensors did not have any measurements. The data acquisition step ends with 47,938 total snapshots. Train, validation, test split (Table \ref{dataset-split}) then was made by snapshot timestamps, and a normalization to map values to $[0-4]$ was applied to all subsets using minimum and maximum values of streamflow and aggregated precipitation observations for the USGS sensor id 05455700 in the training subset. In the end, each snapshot had an input with the size of [8 x 3 x 36] (number of stream gauges x number of time-series sequences used x length of the sequence). The output size similarly was [8 x 36] (Figure \ref{data}). Since a GNN expects graph properties, the graph for sensors was built by considering the hydrological connectivity of the nodes and the distance between USGS sensors as weights.

\begin{figure*}[!ht]
\vskip 0.2in
\begin{center}
\centerline{\includegraphics[height=4cm,keepaspectratio]{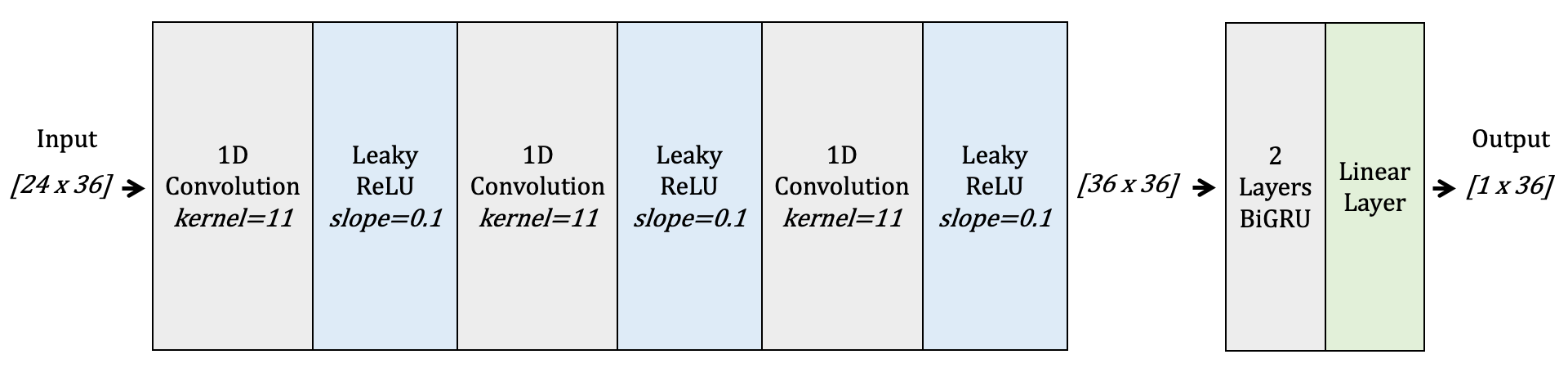}}
\caption{Convolutional Bidirectional GRU network’s architecture.}
\label{gru}
\end{center}
\vskip -0.2in
\end{figure*}
\subsection{Network}

StreamGConvGRU is a GNN model based on Graph Convolutional Gated Recurrent Unit Networks (GConvGRUs) \cite{seo2018structured}. While GConvGRUs have the ability to understand spatial structures with convolutional layers, they also incorporate Gated Recurrent Unit Networks (GRUs) to understand dynamic temporal patterns. For details about GConvGRUs and Graph Convolutional Recurrent Neural Networks, please refer to \cite{seo2018structured}.

The StreamGConvGRU uses three separate GConvGRU subnetworks (Figure \ref{arch})  for each of the three sequences explained in the previous subsection, that is, previous 36 hours of discharge measurements, previous 36 hours of precipitation observations, and next 36 hours of precipitation measurements. The outputs of each of these subnetworks then are summed up and fed to a linear layer that outputs 36 hours of predictions for each of the eight sensors. It should be noted that since a GNN outputs a graph, the product of the network is also a set of sequences, but since the goal is to predict the next discharge levels for the USGS sensor 05455700, the training is done to optimize one sequence while other seven sequences were simply ignored.

We also built various GRU networks for comparison purposes. The best performing GRU network for our dataset was a Convolutional Bidirectional GRU model, which we will refer to as ConvBiGRU (Figure \ref{gru}). ConvBiGRU was trained by feeding a matrix with the shape of [24 x 36] for each snapshot instead of the graph. The matrix was created for each snapshot by combining all sequences for all USGS sensor locations within the river network. The output of the GRU network is only a sequence of subsequent discharge measurements rather than measurements for each of the sensors in the sensor network.

Networks described here were trained using RMSprop \cite{hinton2012neural} optimizer with L1 loss as the cost function on NVIDIA Titan V GPUs. During the training, the best scoring networks’ weights over the validation set were saved, and the results that will be presented in the next section were generated by predicting streamflow for the test dataset. While the ConvBiGRU was implemented using PyTorch numeric computing library \cite{paszke2017automatic}, the StreamGConvGRU was implemented using pytorch-geometric-temporal library \cite{rozemberczki2021pytorch} that is built on top of Pytorch, and pytorch-geometric libraries \cite{fey2019fast}.

\section{Results and Discussions}
Besides the neural network architecture described in the previous section, we define a naive baseline model here. Namely, persistence, which is inherently data-driven, is proposed to be used as a baseline model in streamflow prediction \cite{ghimire2020exploring}.  Persistence is built relying on the idea that “tomorrow will be like today.” It has a straightforward implementation, when prediction is being done at timestep $t$, every future measurement of sensor $s$ for timestep $t’$ is predicted as $s[t-t’]$. In other words, persistence assumes all predictions will be the same as the latest measurement at time $t$.

\begin{figure}
\vskip 0.2in
\begin{center}
\centerline{\includegraphics[width=\columnwidth]{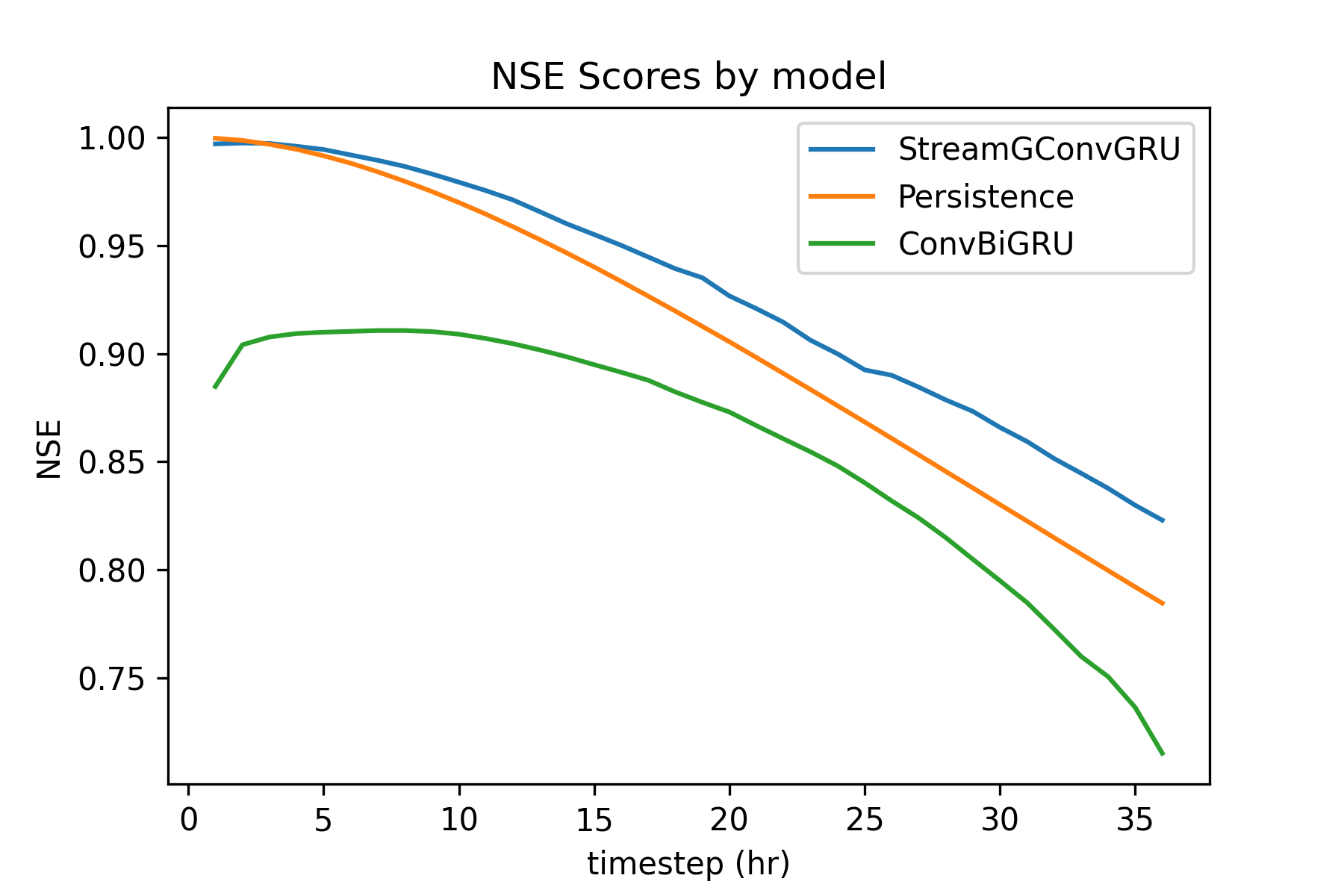}}
\caption{Hourly NSE scores for StreamGConvGRU, persistence and ConvBiGRU.} 
\label{results}
\end{center}
\vskip -0.2in
\end{figure}

To compare persistence, ConvBiGRU, and StreamGConvGRU, we used Nash-Sutcliffe Efficiency (NSE) score \cite{nash1970river}. NSE score is a widely used efficiency metric for streamflow prediction models, both physically based and data-driven applications, and defined as 

\[NSE = 1 - \frac{\sum_{t=1}^{T} (Q_m^t - Q_o^t)^2 }{\sum_{t=1}^{T} (Q_o^t - \overline{Q_o})^2}\]

where $Q_m^t$, $Q_o^t$ and $\overline{Q_o}$ mean modeled discharge at time $t$, observed discharge at time $t$ and mean of observed discharges, respectively. Please note that the best NSE score a streamflow prediction can get is $1.0$. Thus, out of two streamflow prediction models, the model with the NSE score closer to $1.0$ is the better one.

The preliminary results for hourly NSE scores are presented in Figure \ref{results}. As it can be seen in Figure \ref{results}, even though it was the best GRU-based model we built, ConvBiGRU does not get closer to persistence. The StreamGConvGRU model we present in this paper, on the other hand, outperforms both persistence and the ConvBiGRU model by a significant margin after the fifth hour. The performance of the persistence baseline for the first few hours could be explained with the approach it uses. Streamflow rates typically do not change drastically in the first few hours. However, when it starts to change, the StreamGConvGRU model takes the lead as it takes advantage of upstream nodes' data and connectivity. A similar relationship between the proposed model and persistence can be seen in \cite{xiang2020distributed}. Also, we want to stress that various others RNNs we built to compare with StreamGConvGRU did not produce comparable NSE scores; consequently, we decided not to include them here for the sake of simplicity.

\section{Conclusions}
The importance of streamflow and, consequently, flood prediction increases as a result of the devastating effects of climate change. This paper presented an approach where we employed a Graph Convolutional GRU Networks based model, StreamGConvGRU, to predict 36 hours of streamflow for a sensor location using the upstream river network. As shown in the preliminary results, the StreamGConvGRU provides better performance than the persistence baseline and a Convolutional Bidirectional GRU model in our study region for short-term hourly streamflow prediction. For the future work, we aim to focus on the following three points: (1) exploring GConvGRUs’ abilities with a more extensive river network; (2) prediction of greater lead time on streamflow; (3) prediction of streamflow on each node in the river network instead of focusing only on one node at the outlet. 




\bibliography{main}
\bibliographystyle{icml2021}

\end{document}